\documentclass[sigconf,10pt]{acmart}
\usepackage{bbm}
\usepackage{microtype}
\usepackage[caption=false,font=footnotesize]{subfig}
\usepackage[linesnumbered, ruled, noend]{algorithm2e}
\AtBeginDocument{%
  }

\setcopyright{acmlicensed}
\copyrightyear{2018}
\acmYear{2018}
\acmDOI{XXXXXXX.XXXXXXX}

\acmConference[Conference acronym 'XX]{Make sure to enter the correct
  conference title from your rights confirmation emai}{June 03--05,
  2018}{Woodstock, NY}

\usepackage{bbm}

\begin{document}

\title{FLAD: Federated Learning for LLM-based Autonomous Driving in Vehicle-Edge-Cloud Networks}


\author{Tianao Xiang}
\affiliation{
  \institution{Northeastern University}
  \country{China}
}
\email{2010652@stu.neu.edu.cn}

\author{Mingjian Zhi}
\affiliation{
  \institution{Northeastern University}
  \country{China}
}
\email{2110657@stu.neu.edu.cn}

\author{Yuanguo Bi}
\affiliation{
  \institution{Northeastern University}
  \country{China}
}
\email{biyuanguo@mail.neu.edu.cn}

\author{Lin Cai}
\affiliation{
  \institution{University of Victoria}
  \country{Canada}
}
\email{cai@ece.uvic.ca}

\author{Yuhao Chen}
\affiliation{
  \institution{University of Victoria}
  \country{Canada}
}
\email{yuhaoc@uvic.ca}
\renewcommand{\shortauthors}{Xiang et al.}

\begin{abstract}
  Large Language Models (LLMs) have impressive data fusion and reasoning capabilities for autonomous driving (AD). 
  However, training LLMs for AD faces significant challenges including high computation transmission costs, and privacy concerns associated with sensitive driving data. 
  Federated Learning (FL) is promising for enabling autonomous vehicles (AVs) to collaboratively train models without sharing raw data. 
  We present Federated LLM-based Autonomous Driving (FLAD), an FL framework that leverages distributed multimodal sensory data across AVs in heterogeneous environment. 
  FLAD has three key innovations: (1) a cloud-edge-vehicle collaborative architecture that reduces communication delay and preserving data privacy; (2) an intelligent parallelized collaborative training with a communication scheduling mechanism that optimizes training efficiency, leveraging end-devices otherwise having insufficient resources for model training; and (3) a knowledge distillation method that personalizes LLM according to heterogeneous edge data. 
  In addition, we prototype FLAD in a testbed with NVIDIA Jetsons, overcoming practical implementation challenges including CPU/GPU memory sharing in resource-constrained devices, dynamic model partitions, and fault-tolerant training.
  Extensive experimental evaluation demonstrates that FLAD achieves superior end-to-end AD performance while efficiently utilizing distributed vehicular resources, opening up new possibilities for future collaborative AD model training and knowledge sharing.
\end{abstract}

\ccsdesc[500]{Human-centered computing~Mobile computing}
\ccsdesc[500]{Computing methodologies~Distributed artificial intelligence}

\keywords{Autonomous driving, autonomous vehicle, distributed machine learning, federated learning, hybrid parallelism}

\received{20 February 2007}
\received[revised]{12 March 2009}
\received[accepted]{5 June 2009}

\maketitle
\section{Introduction}
Large Language Models (LLMs) offer unprecedented capabilities in data fusion, reasoning, and decision-making, making them a promising candidate for end-to-end autonomous driving solution~\cite{dong2024generalizing,wang2024drivedreamer,chen2024driving}. 
However, current LLM-based AD systems heavily rely on centralized data centers with extensive domain-specific datasets, raising significant concerns about data privacy, communication overhead, and real-time performance. While existing AD research has focused on centralized model training and inference, the increasing demand for privacy-preserving and personalization in autonomous vehicles (AVs) necessitates a shift toward distributed learning approaches~\cite{chellapandi2023survey,sirohi2023federated}. 
\par
In this context, Federated Learning (FL) emerges as a compelling solution, enabling collaborative model training while preserving data locality, complemented by knowledge distillation techniques that adapt large models to resource constrained edge devices~\cite{gou2021knowledge,kairouz2021advances}.
Future AVs will come in a wide range of sizes, from large buses and trucks to compact drones, catering to diverse transportation and exploration needs. Nearby AVs can work together to further personalize the model to improve model performance, considering the heterogeneity of vehicles and environment in different locations~\cite{nextmsc}. 
\par
Achieving efficient on-vehicle learning for AD tasks presents significant challenges in both computational efficiency and resource management~\cite{yang2021edge,prathiba2021federated}. 
Recent research shows that training compact models on mobile devices can be up to 160$\times$ slower than on GPU servers, leading to the gap between computational demands and the limited processing power of onboard hardware~\cite{ye2024asteroid,hao2021eddl}. 
This challenge becomes more significant when dealing with multimodal sensor data, which requires increasingly complex model architectures for real-time decision-making. 
While various optimization strategies—such as model compression, pruning, and efficient architecture design—have been explored, the fundamental resource constraints of on-board systems still hinder effective deployment.
\par
We observe that vehicles operating at network edges can not only access private driving data, but also represent an underutilized distributed computing potential~\cite{manias2021making,chellapandi2023federated}. 
Inspired by this observation, our approach uses nearby vehicles as a collaborative resource pool, enabling accelerated edge-based model training while preserving privacy.
\par
Distributed training for AD presents unique challenges not encountered in traditional data center environments. Vehicles operate under strict resource constraints, experience high network instability due to mobility, and deal with private heterogeneous data from different types of sensors and environment. 
\par
To address these challenges, we present Federated LLM-based Autonomous Driving (FLAD), a cloud-edge-vehicle collaborative system that orchestrates model training while ensuring fault tolerance and privacy preservation. FLAD introduces two key innovations that directly tackle the identified bottlenecks: (1) Federated Hybrid Data Parallelism (FHDP), which combines FL with pipeline parallelism to maximize resource utilization while preserving data privacy, enabling collaborative intelligence without exposing sensitive driving data; (2) A cloud-edge-vehicle collaborative architecture that leverages LLM reasoning capabilities for AD tasks while distributing computational workloads across three layers, effectively addressing the resource constraints of vehicular platforms.
\par
We implement FLAD in a realistic testbeds comprising multiple heterogeneous edge devices. Extensive evaluations show that FLAD achieves 70\% throughput of centralized training approaches while maintaining robust performance under dynamic mobility conditions and network fluctuations. Additionally, FLAD demonstrates robust adaptation to device failures, achieving recovery 5 $\times$ faster than baseline approaches while maintaining throughput stability.
\par
The main contributions are summarized as follows.

    \textbf{(1) We propose FLAD, a novel cloud-edge-vehicle collaborative framework that enables efficient deployment of LLM-based autonomous driving while preserving data privacy.} FLAD addresses the tension between computational requirements and resource constraints by facilitating training with pipeline parallelism, creating a distributed intelligence framework for vehicular environments.
    
    \textbf{(2) We develop SWIFT (Speedy Weight-based Intelligent Fast Two-phase scheduler),} which schedules resource-constrained vehicles to participate in distributed training. Combining stability-based vehicle ordering with DQN optimization, SWIFT generates mobility-aware pipeline configurations that significantly enhance performance in dynamic and heterogeneous vehicular environments.
    
    \textbf{(3) We design a resilient quick recovery mechanism featuring preventive pipeline template generation and edge-aided backup strategies.} This dual-module approach enables FLAD to recover quickly from network failures due to high mobility, while minimizing communication overhead through the design of different model partitioning solutions, ensuring training continuity despite the inherent instability of vehicle networks.
    
    \textbf{(4) We implement and evaluate FLAD on real-world testbeds comprising heterogeneous terminal devices,} overcoming practical implementation challenges including CPU/GPU memory sharing in resource-constrained devices, dynamic model partitions, and fault-tolerant training. The test result demonstrate significant improvements in both vision tasks and AD metrics. Specifically, FLAD achieves 70\% throughput of that with centralized training while maintaining robust performance under dynamic mobility conditions and network fluctuations.
\par
The rest of the paper is organized as follows. The motivation and design goals are presented in Sec. 2, and an overview of the FLAD framework is given in Sec. 3. The main design of federated hybrid data parallelism is presented in Sec. 4, followed by implementation details in Sec. 5. Sec. 6 presents the evaluation results, and Sec. 7 presents related work, followed by concluding remarks and future work in Sec. 8.  
\section{Motivations}
We first investigate the challenges and design objectives for FLAD, where four critical challenges are identified: unstable vehicle availability, wireless resource contention, unbalanced computational capabilities, and severe memory constraints. 

\subsection{Unstable Available Client Set}
Vehicular federated learning (VFL) suffers from inherently unstable client availability due to vehicle mobility. 
Unlike traditional FL settings where clients remain connected throughout training rounds, vehicles may leave network coverage before finishing a full FL round. 
Moreover, the available vehicles to participate in training change unpredictably as vehicles enter and depart network coverage areas, making it diffcult to maintain consistent training cohorts between rounds. 
This instability introduces training discontinuities that directly impede model convergence and significantly degrade learning performance in automotive deployments.
\subsection{
Competitive Model Transmission
}
In VFL, the assumption that model transmission time is negligible no longer holds due to constrained communication resources. 
Data exchange in vehicular networks over wireless links may suffer from severe bandwidth limitations caused by concurrent transmission demands and complex communication messages defined in the current protocols. 
These constraints are exacerbated by dynamic wireless link quality, due to Doppler shift, multipath fading, shadowing, and heterogeneous transmission requirements.
Vehicles must compete for limited bandwidth, causing unpredictable model transmission delays and reduced training efficiency.

\subsection{Unbalanced Local Training Workload}
Heterogeneous vehicle capabilities can cause significant training workload imbalances in the VFL systems. 
Vehicles have widely varying data volumes, computational resources, and network dwell time. 
Consequently, applying uniform training settings (such as identical local epoch counts) across all clients leads to inefficient resource utilization and training bottlenecks. 
For example, a vehicle with substantial computational capacity and high-quality data can efficiently complete multiple local epochs to contribute valuable model updates. 
Conversely, requiring the same epoch count from vehicles with limited resources not only wastes computational cycles on potentially lower-quality updates but may also cause these resource-constrained vehicles to exceed their available dwell time before completing training.

\subsection{
Hardware Memory Constraints
}
Current AD systems face significant memory requirements that limit on-vehicle deployment. 
As the training systems evolve, the memory requirements have grown explosively from approximately 2GB for CNN-based perception models to over 16GB for transformer-based architectures. 
This growth presents significant implementation challenges for vehicles with fixed hardwares.
\par
Transformer-based architectures, while delivering superior performance for AD tasks, impose considerable memory demand, primarily due to their attention mechanisms. 
A typical multi-head attention layer in a vision transformer requires storing $\mathcal{O}(n^2)$ attention maps for $n$ input tokens, translating to hundreds of megabytes for a single forward pass with high-resolution sensor data. 
This memory intensity is a significant barrier to both deployment and critical fine-tuning operations on vehicle platforms, where available memory can be less than 12GB.
\par
In addition, the AD model must have sufficient perception and planning capabilities. Otherwise, model performance will be degraded, which may lead to traffic accidents due to inaccurate perception and decision making. 
Such huge model parameters demonstrate the urgent need for memory-efficient methods that enable sophisticated model adaptation in hardware constrained AVs.
\subsection{Opportunities of LLM-based AD
}
The challenges in VFL present substantial barriers for AD applications, while recent LLM advancements offer compelling solutions to address some of them. 
LLMs demonstrate multimodal reasoning capabilities, seamlessly integrating diverse sensor inputs with contextual knowledge to interpret complex traffic scenarios with unprecedented comprehensiveness.
Parameter-efficient fine-tuning techniques such as Low-Rank Adaptation significantly reduce LLM adaptation memory requirements by modifying only 0.1-1\% of parameters, which makes on-vehicle model personalization feasible even with local memory constraints. 
Furthermore, combined with FL principles, LLMs create an ideal framework for privacy preservation by processing high-level semantic representations rather than raw sensor data. This approach maintains data locality while enabling collaborative model improvement across heterogeneous vehicle fleets.
These complementary capabilities motivate our development of FLAD, which orchestrates FL across cloud, edge, and vehicle layers to enable efficient, privacy-preserving AD systems that overcome the limitations inherent in traditional centralized approaches.

\section{Framework Overview}
As illustrated in Fig.~\ref{fig:entire architecture}, FLAD is a three-layer architecture including vehicle layer, edge layer, and cloud layer, each serving distinct roles in AD model training.
\subsection{Training procedure of FLAD}
\begin{figure}[!t]
    \centering
    \includegraphics[width=0.9\linewidth]{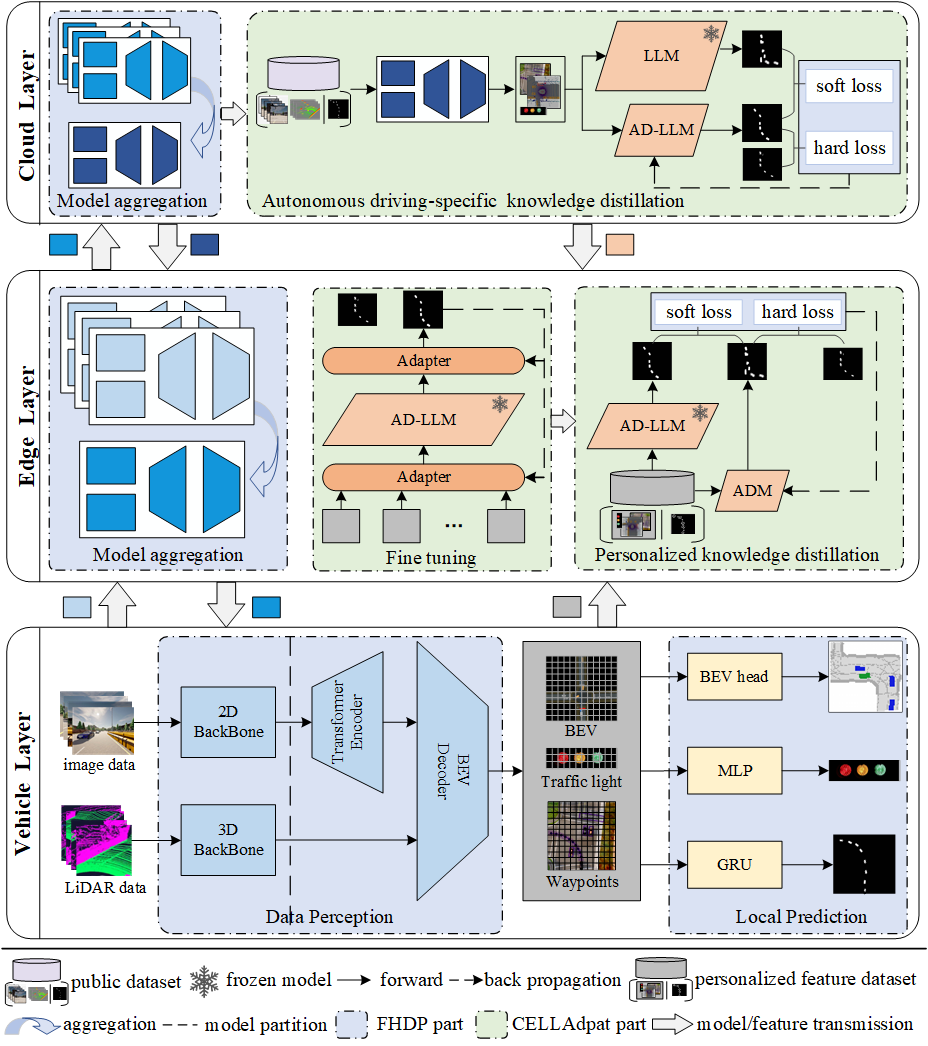}
    \caption{Training procedure in FLAD.}
    \label{fig:entire architecture}
\end{figure}

FLAD's training process consists of two key components, i.e., a vision encoder to process sensed image and LiDAR data by each AV and an LLM-based AD decision model to process multimodal inputs, including road maps, traffic signs, and all types of contextual information. 
As shown in Fig.~\ref{fig:entire architecture}, the vision encoder is trained with an FL paradigm among the three layers using local sensory data by each AV, i.e., the vision encoder models trained by vehicles are first aggregated at the edge server, then further aggregated at the cloud server; the resulting aggregated model is shared with all vehicles for the next round of training, and this process continues iteratively.
The AD domain-specific LLM is distilled using public AD datasets in the cloud and transmitted to the edge layer, which is AD-LLM.
The AD-LLM is fine-tuned with features extracted by vision encoders of vehicles. 
To optimize the inference speed, the edge layer performs knowledge distillation to create a compact Autonomous Driving Model (ADM).
The specific function of each layer in training procedure of FLAD is demonstrated as follows. 
\par
The vehicle layer processes multimodal sensor data locally and participates for critical perception tasks including Bird's Eye View (BEV) perception, traffic light recognition, and waypoint prediction to improve the feature extraction capability in AD tasks of vision encoder. 
\par
The edge layer serves as a regional intelligence hub with dual functions: (1) aggregating trained vision encoder from participating vehicles to improve perception capability, and (2) maintaining an AD-LLM that makes driving decisions based on the aggregated perceptual knowledge. 
\par
The cloud layer coordinates knowledge across geographic regions by aggregating models from edge layers and orchestrating knowledge transfer from general LLMs to specialized AD-LLMs through distillation. 

\subsection{Inference procedure of FLAD}
\begin{figure}[!t]
    \centering
    \includegraphics[width=0.9\linewidth]{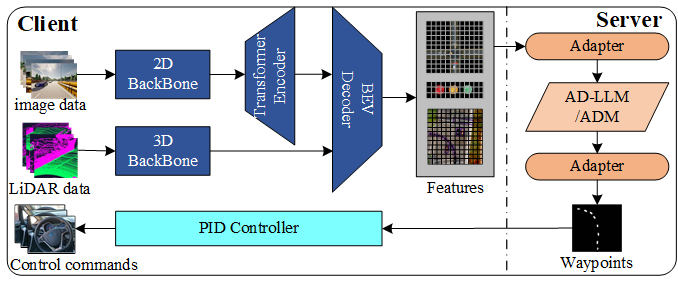}
    \caption{Inference procedure in FLAD.}
    \label{fig:entire architecture inference}
\end{figure}

The FLAD inference workflow, illustrated in Fig.~\ref{fig:entire architecture inference}, distributes computational burdens across the vehicle-edge hierarchy to optimize decision-making efficiency. 
Vehicles capture multimodal sensory data and process them through the FL trained vision encoder to extract compact semantic features, reducing data transmission volume while preserving privacy. 
These high-level features are transmitted to the edge-hosted AD-LLM, which integrates perceptual information with contextual knowledge (e.g., navigation and notice instruction) to generate driving decisions. 
The AD-LLM performs sequential reasoning over the received features to produce future waypoints, which are returned to the vehicle. 
Then the waypoints are transformed to vehicle control commands through the PID controller in each vehicle.
\par
In summary, with the three-layer design, the cloud can fully utilize the knowledge of a large number of AVs to generate and fine-tune generalized AD-LLM, thanks to the rich resources in the cloud. The edge can use personalized and distilled AD-LLMs to assist AVs in making complicated AD decisions promptly, such as waypoint generation, considering all contextual information and real-time encoded sensing data. The personalized AD-LLM in the edge is tailored to the area, so they can be more accurate in guiding the vehicles nearby. AVs with limited computing and memory resources can leverage the AD-LLM models in the edge for intelligent AD decisions, and they can also contribute to the training process without compromising their data privacy.
\subsection{Challenges in implementing FLAD}
Although promising, implementing this three-layer architecture presents three challenges: (1) vehicle mobility brings inherently unstable participant availability, disrupting FL training continuity; (2) computational heterogeneity between vehicles requires adaptive workload allocation mechanisms; and (3) the memory-intensive requirements of both LLMs and perception models for deployment on resource-constrained vehicle platforms.
\par
To solve these challenges, we design two primary workflows:
(1) Federated Hybrid Data Parallelism among edge-vehicle (FHDP): Implements FL between vehicles and edge servers to train vision encoders using distributed vehicle data while preserving privacy and reducing communication overhead.
(2) Cloud-Edge LLM Adaptation~(CELLAdapt): Combines fine-tuning and knowledge distillation to enhance edge server performance. Fine-tuning adapts the LLM to specific edge server contexts, while knowledge distillation efficiently transfers driving-relevant capabilities from comprehensive cloud LLMs to more compact edge AD-LLMs.

\section{Federated Hybrid Data Parallelism}
\begin{figure}
    \centering
    \includegraphics[width=\linewidth]{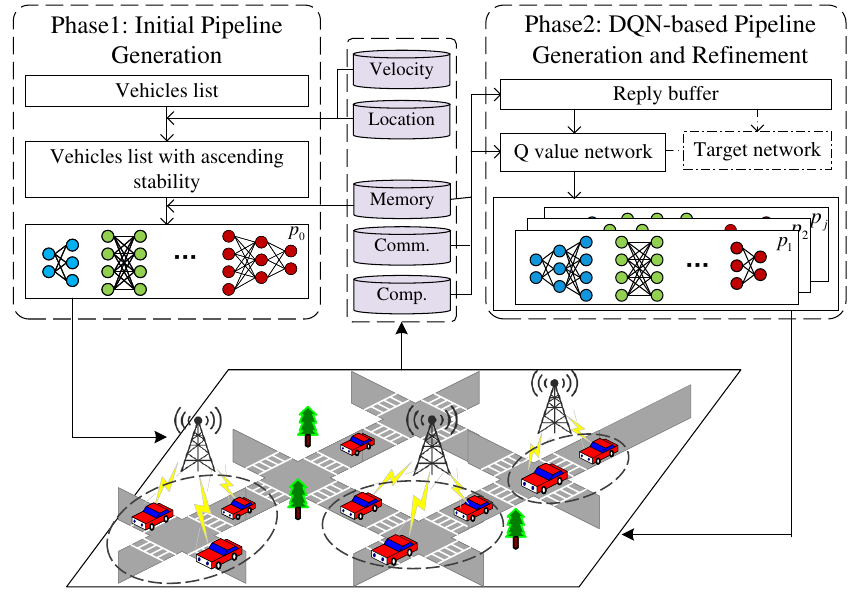}
    \caption{The overview of SWIFT in FHDP.}
    \label{fig:swift architecture}
\end{figure}
AD tasks require vehicles to extract representations from sensory data using locally trained vision encoders, which then enable AD-LLM at edge to make driving decisions. 
However, many vehicles lack sufficient computation, memory, and communication resources to independently train their encoders, which needs collaborative training approaches.
\par
Traditional HDP, which distributes model partitions across clusters of devices, faces significant limitations in vehicle environments due to data isolation and heterogeneity. 
To overcome these limitations, we propose Federated Hybrid Data Parallelism (FHDP), which uniquely leverages distributed data across all participating vehicles to improve model performance without additional computational overhead.
\par
FHDP distributes model partitions sequentially across vehicle stages, creating computational pipelines. Each stage processes a specific model partition before forwarding results. Since pipeline interruptions due to vehicle disconnections force costly reconfigurations that delay convergence, our system employs intelligent clustering based on resource availability, mobility predictions, communication latency, and computational capabilities to maintain optimal performance in dynamic vehicular environments.
\par
Traditional HDP architectures use static clusters where only first-stage devices contribute their local data, while other devices provide computational resources without data input. The inter-cluster organization follows a data parallelism approach, with clusters functioning as work nodes that train models using exclusively first-stage data.
\par
This conventional design is incompatible with FLAD due to inefficient data utilization. Vehicle data's inherent non-i.i.d. characteristics create heterogeneous distributions across both clusters and individual vehicles. Limiting training to first-stage data significantly reduces model adaptability across the fleet, degrading overall performance.
\par
Our proposed FHDP overcomes these limitations through a dynamic stage exchange mechanism. Unlike traditional HDP, FHDP enables every vehicle to contribute its local data regardless of pipeline position. Vehicles systematically rotate through pipeline stages, allowing each participant to serve as a data provider during different training iterations. This approach maximizes data utilization across all participating vehicles while preserving the computational efficiency benefits of pipeline parallelism.
\par
As shown in Fig~\ref{fig:swift architecture}, FHDP implements a two-stage resource management approach that combines long-term strategic planning with dynamic adaptation mechanisms.
The first stage focuses on long-term static planning through two key components. 
Initially, it performs a vehicle availability assessment, considering both current and predicted vehicle trajectory within the area. 
Furthermore, an intelligent clustering process is proposed that jointly optimizes model capacity and geographical distribution. 
In second stage, the framework employs a Double Q Network (DQN)-based scheduler for efficient inter-cluster resource scheduling, culminating in the design of pipeline execution templates that optimize intra-cluster resource utilization.
The second stage addresses the dynamic nature of vehicular networks through short-term adaptation mechanisms. We presnet a dynamic quick recovery mechanism for short-term reliability for network topology changes, complemented by an RSU-aided backup system that ensures training continuity during network perturbations.
\subsection{Static Planning for Stability Training}
The static planning phase establishes a framework for extended training periods by generating stable vehicle clusters and corresponding pipeline parallelism templates, ensuring consistent and efficient training processes.
\subsubsection{\textbf{Analysis of Vehicular Availability Status}}
Given the inherent mobility of vehicular networks, comprehensive availability assessment becomes crucial for maintaining training stability. Our framework evaluates vehicle eligibility through three key metrics: dwell time, computational capacity, and memory availability.
Dwell time prediction utilizes historical edge server data, with samples represented as $\mathbf{dwl}=[dwl_1,dwl_2,\ldots] \in \mathbb{R}_+$, where $dwl_i=[arv_i,dep_i]$ captures arrival-departure intervals. For trajectory data $\mathbf{B} = [\mathbf{b}_1,\ldots,\mathbf{b}_N]$, we predict dwell times for unseen routes $\mathbf{b}q \notin \mathbf{B}$ by formulating a MAPE regression problem: $\underset{R}{\min}\sum_{i=1}^{N}\frac{\vert a_i - R(\mathbf{b}_i) \vert}{a_i} + \Omega(R)$, where $R(\mathbf{b}_i)$ estimates route sojourn time and $\Omega(R)$ provides regularization. We solve this optimization using a wide-deep-recurrent learning architecture~\cite{Wang2018Learning}.
\par
Computational capacity is quantified through GPU performance, measured in TFLOPS ($cmp_i$), obtained directly from hardware specifications. 
\par
Memory availability is assessed through two ways: dedicated graphic memory size for systems with discrete GPUs, or $Mem_{limit}-Mem_{sys}$ for unified memory systems, where $Mem_{limit}$ represents physical memory capacity and $Mem_{sys}$ denotes peak system process usage. 
The relationship between computation power and dwell time is characterized by:
\begin{equation}
dwl_i \cdot cmp_i \leq \alpha M_{cmp} e_{req},\label{eq: computation power conditions}    
\end{equation}
where $M_{cmp}$ represents the computational volume per epoch, $e_{req}$ specifies the minimum required epochs, and $\alpha \in [0,1]$ defines the minimum proportion of the model training task to be completed.
\par
After evaluating the availability of vehicles, the vehicles that have enough resources to training models locally need to be filtered because these vehicles do not need to collaborate with other vehicles to train models.
The certification of enough resource vehicles is conditioned on
\begin{equation}
    \begin{aligned}
        dwl_i \cdot cmp_i \geq M_{cmp} e_{req};~mem_i \geq M_{cap},        
    \end{aligned}\label{eq: condition of powerful vehicle}
\end{equation}
where $mem_i$ is the available memory and $M_{cap}$ is the required memory footprint of model $M$ including the activation, gradient, and optimizer status which is usually $10\times$ the model size.
According to Eq.~\eqref{eq: condition of powerful vehicle}, available vehicles can be categorized into two kinds: resource-sufficient vehicles $n^{rs}_i$ and resource-limited vehicles $n^{rl}_i$.
\subsubsection{\textbf{Model capacity and geography-based clustering}}
To optimize the utilization of resource-limited vehicles, denoted as $v \in \mathcal{V}_{rl}$, we propose a clustering algorithm that considers both model capacity and reliability metrics. 
This algorithm aggregates resource-limited vehicles into clusters capable of collaborative model training. 
The clustering approach ensures that vehicles with insufficient individual resources can collectively form a computational unit with adequate capacity to execute the training process.
\par
For the formation of a stable and resource-sufficient cluster, we first propose a stability model for vehicle mobility.
We model the area as a grid of unit cells $C = \{c_1,c_2, \cdots c_{|C|}\}$ where $|C| = {R^2}/{\varrho^2}$, with $R$ being the area range and $\varrho$ the unit distance. 
Each vehicle $v$ has its communication radius $R_v$ covering cells $C_v \subset C$ where $|C_v| = {\pi R^2_v}/{\varrho^2}$, defining its neighbor set $\mathcal{NB}_v$. 
The distance between vehicles is measured as $n(c_v \rightarrow c_{nb})$, representing the cell-count between positions.
Vehicle mobility is modeled using Discrete Time Markov Chains (DTMC), represented as patterns $\mathbf{Mob} = \{m_1,m_2,\cdots,m_K\}$ derived from historical data. 
Each pattern $m_k$ has associated transition probabilities $P(c_i \rightarrow c_j|m_k)$ between cells. 
Future position prediction for vehicle $v$ is formulated as
\begin{equation}
   P(c_{n}\rightarrow c_f|H) = \sum_{m_a\in \mathbf{Mob}} P(m_a|H) P(c_{n}\rightarrow c_f|m_a),
\end{equation}
where $H$ represents historical trajectory, $c_n$ is current cell, and $c_f$ is future cell.
The joint probability of vehicles $v$ and $nb$ occupying specific cells at time $t$ is
\begin{equation}
   P(c^{nb}_f(t),c^v_f(t)) = P(c^{nb}_{n}\rightarrow c^{nb}_f|H_{nb})P(c^{v}_{n}\rightarrow c^{v}_f|H_v).
\end{equation}
We quantify neighbor stability as
\begin{equation}
   Stb_{nb} = \sum_{t=0}^{dwl_v} \sum_{c^v_f(t) \in C} RD_{nb}(t),
\end{equation}
where relative distance $RD_{nb}(t)$ is the distance between $v$ and $nb$ at time $t$, which accounts for the predicted changes in position.
Resource capability is quantified as $mem_{nb} = \beta M_{cap}$ with $\beta \in (0,1]$ representing the memory-to-model capacity ratio.
The goal is to form a cluster that has sufficient resources to collaborate and is long-term stable, which can be formulated as
\begin{eqnarray}
    \begin{aligned}
        &\arg \underset{Clu_v \in C_v}{\max} \sum_{nb \in Clu_v} Stb_{nb} \\
        &\text{c1:~}\underset{Clu_v \in C_v}{\sum} mem_{nb} > M_{cap},~\forall nb \in Clu_v\\
        &\text{c2:~}\underset{Clu_v \in C_v}{\sum} dwl_{nb} com_{nb} > e \alpha' M_{cmp}\\
        &\text{c3:~}\vert Clu_v \vert \leq \vert C_v(t) \vert,\ \forall t \leq dwl_{v},
    \end{aligned}\label{eqn: cluster formation problem}
\end{eqnarray}
where $e$ is the number of epochs that cluster can stably execute, $\alpha'\geq 1$ is a redundancy parameter to provide fault tolerance for potential failures, and $C_v(t)$ denotes the neighbor set of $v$ at time $t$. 
The objective function of \eqref{eqn: cluster formation problem} includes three parts: the stability of the cluster, the epochs that can execute, and the cluster size.
The former two aim to maximize the training stability and large cluster size is penalized to reduce the sudden change risk.
With the clustering process, the resource-sufficient cluster can be considered as a single resource-sufficient client for FL training.
\subsubsection{\textbf{Pipeline template design for intra-cluster scheduling}}
The entire FHDP is a hybrid parallelism scheme, including a pipeline parallelism-based method to improve the resource utilization of resources-limited vehicles in a cluster, and a data parallelism based-method among the resource-sufficient clients, including resource sufficient vehicles and clusters.
Vehicle mobility induces temporal variations in intra-cluster stability, requiring adaptive pipeline configurations to maintain system robustness.
Intra-cluster pipeline templates provide a robust execution framework that mitigates unpredictable changes in cluster membership caused by vehicle mobility.
Based on each vehicle's available memory $mem_v$, computational capability $com_{v}$, and the structure of the vision encoder, we can accurately estimate the capacity of model partition that each vehicle can support.
\par
{\textbf{Complexity of Vision Encoder.}}
The vision encoder employs a DAG architecture with four components: ResNet-based RGB backbone, PointPillarNet-based LiDAR backbone, transformer encoder for feature extraction, and transformer decoders for multimodal fusion. We perform detailed component level FLOPs analysis to establish accurate computational profiles, enabling efficient workload distribution across heterogeneous vehicular resources with varying processing capabilities.
\par
Specifically, for systematic model partitioning, we represent the vision encoder as a DAG where nodes correspond to specific modules (Conv, MaxPool, Attention, etc.) and edges encode data dependencies between these modules.
To generate sequential execution stages, we apply topological sorting to transform this graph representation into an ordered layer sequence optimized for distributed processing.
Let $M^{RGB}_{cmp}$, $M^{Lid}_{cmp}$, $M^{Enc}_{cmp}$ and $M^{Dnc}_{cmp}$ denote the computation of RGB, Lidar backbone, encoder, and BEV decoder modules, respectively.
The module computation per epoch can be formulated as
\begin{equation}
    M_{cmp} = M^{*}_{cmp} N_{batch},
    \label{eq:computation cost}
\end{equation}
where $M_{cmp}$ is the model FLOPS per sample, and $N_{batch}$ is the number of batch per epoch. $*$ represents $RGB$ or $Lid$ or $Enc$ or $Dnc$ for different model partitioning.
The computation time of the module of vehicle $v$ is 
\begin{equation}
    t_{cmp} = {M_{cmp}\nu}/{(cmp_v\mu)},\label{eq:training time}
\end{equation}
where $\mu \in [0.3,0.7]$ is the utilization of GPU performance and $\nu \in [1.1,1.5]$ is the cost of memory bandwidth overhead moving data between memory and processors.
Let $com_v$ denote the communication capability of $v$, and the communication time of $M^v_{cap}$ in per epoch can be formulated as
\begin{equation}
    t_{com} = {2M_{cap}N_{batch}\nu}/{com_v} ,\label{eq:communication time}
\end{equation}
where $2M_{cap}$ is the communication volume in the forward and backward processes.
\par
Let $M^{u}_{cap}$ denote the capacity of a unit model partition, which is determined by the vehicle with least resources.
The entire model can be split into multiple partitions as $M_{cap}=\sum^{K}_{k=1} M^{u,k}_{cap}$.
Due to the different partition architecture, each $M^{u,k}_{cap}$ has different computational volume $M^{u,k}_{cmp}$ and communication volume $M^{u,k}_{com}$.
\par
Let $Clu$ denote an FLAD cluster, we define a DAG $G_{Clu} = (V,E)$ to present a pipeline template of $Clu$, where $V$ is the set of vertices (vehicles) and $E$ is the set of edges denoted execution dependencies~\cite{stynes2016probabilistic}.
\par
Let $|V|$ denote the number of vehicles in the cluster. Each vehicle $v \in V$ has two critical attributes: available memory $mem_v$ and computational capability $cmp_v$. The directed edge set $E$ contains ordered pairs $(u,v)$ where $u,v \in V$, representing execution dependencies. Specifically, directed edge $(u,v)$ enforces that vehicle $u$ must complete its computation before vehicle $v$ begins processing, establishing a directed acyclic structure that ensures sequential execution and prevents pipeline deadlocks.
\par
We denote a path $p=(v_1,v_2,\cdots,v_{|V|})$ in $G_{Clu}$ as a pipeline execution order, where $v_i\neq v_j\in V$.
In addition, we use $\mathcal{P}_{Clu} = \{M^v_{cap}|v\in V\}$ to represent a model partition strategy of $Clu$, where $M^v_{cap}=\sum^{K_v}_{k=1}M^{u,k}_{cap}$ and $\sum_{M^v_{cap}\in \mathcal{P}_{Clu}} M^v_{cap} = M_{cap}$. 
Given Eqs.~\eqref{eq:training time} and \eqref{eq:communication time}, the execution time of path $p$ under a model partition strategy $\mathcal{P}$ can be formulated as 
\begin{equation}
    t_{path}(p,\mathcal{P}) = \sum_{v\in p} t^{v}_{cmp} + \sum_{v\in p\setminus v_{|V|}} t^{v}_{com}, \label{eq: path time}
\end{equation}
where $t^{v}_{cmp}$ and $t^{v}_{com}$ is the training and communication time of vehicle $v$ with model partition $M_v$.
\par
Therefore, the pipeline generation problem in an FLAD cluster $Clu$ can be formulated as 
\begin{eqnarray}
    \begin{aligned}
        &\underset{p,\mathcal{P}}{\arg \min}~t_{path}(p,\mathcal{P})& \label{eq:objective}\\
        &\text{c1:~}\sum_{M^v_{cap}\in \mathcal{P}_{Clu}} M^v_{cap} = M_{cap}& \label{eq:constraint1}\\
        &\text{c2:~}M^v_{cap} \leq mem_v,\forall v \in V& \label{eq:constraint2}\\
        &\text{c3:~}\forall (u,v) \in E:~\text{start}_v \geq \text{end}_u + t^u_{com}& \label{eq:constraint3}\\
        &\text{c4:~}p=(v_1,v_2,\cdots,v{|V|}),v_i\neq v_j \in V& \label{eq:constraint4}\\
        &\text{c5:~}M^v_{cap}\cap M^u_{cap}=\phi,\forall M^v_{cap}, M^u_{cap}\in \mathcal{P}_{Clu}& \label{eq:constraint5}
    \end{aligned}\label{eq:pipeline generation}
\end{eqnarray}
Eq.~\eqref{eq:objective} aims to minimize total path execution time while satisfying five key constraints: \textit{c1} ensures complete model partitioning, \textit{c2} enforces per-vehicle memory limits, \textit{c3} maintains DAG precedence relationships, \textit{c4} requires valid non-repeating vehicle paths, and \textit{c5} ensures disjoint model partitions. This approach simultaneously optimizes both partitioning strategies and execution ordering while respecting all resource and operational constraints in the FLAD cluster.
\par
{\textbf{Design of SWIFT.}}
The pipeline generation problem in Eq.~\eqref{eq:pipeline generation} combines graph partitioning, scheduling, and resource allocation, making it NP-hard. 
While existing heuristic and learning-based methods can find approximate solutions, their lengthy computation time poses a critical challenge in dynamic FLAD clusters, where vehicle mobility can render solutions obsolete before implementation.
\par
To address this challenge, we propose SWIFT, which exploits a fundamental VFL requirement: the necessity to process data from all participating vehicles. 
This requirement establishes a natural constraint of maintaining $|V|$ essential pipelines for each vehicle. 
\par
SWIFT leverages this insight to decompose pipeline generation into two phases: establishing an initial pipeline for quick start and intelligent pipeline generation and refinement for remaining vehicles. 
This structured approach significantly reduces the start-up latency and computational complexity compared to simultaneous optimization of all possible configurations.
\par
{\textbf{Initial Pipeline Generation.}} In the first phase, SWIFT establishes an initial pipeline using greedy matching based on cluster stability scores $Stb_v$. 
These scores quantify each vehicle's likelihood of maintaining cluster presence, crucial since early pipeline stages require longer vehicle participation. The algorithm assigns model partitions $M^{v}_{cap}$ through greedy matching, with each vehicle receiving the maximum partition size that satisfies its memory constraint $mem_i$.
\par
{\textbf{DQN-based Pipeline Generation.}} In the second phase, for subsequent pipelines, SWIFT employs a DQN-based approach that builds on the stability-prioritized sequence. 
Specifically, let $v_i$ denote the first stage of the initial pipeline.  
For the remaining vehicles in set $V' = V\setminus {v_i}$, we arrange them in ascending order of their stability scores. 
The pipeline generation process then proceeds iteratively: we select the vehicle $v_j$ with the lowest stability score from $V'$ as the first stage of a new pipeline. 
Using our DQN approach, we generate an optimal pipeline path $p_j=(v_j,v_{j1},\ldots,v_{jk})$ and its corresponding model partition strategy. 
This process continues until each vehicle in $V$ serves as the starting point for at least one pipeline, ensuring comprehensive data utilization across the cluster.
\par
The DQN state space for pipeline generation encompasses five key components that capture the system's complete configuration: (1) available model capacity, tracked as $(M_{cap} - \sum M^v_{cap})$; (2) current model partitions, represented by vehicle-partition pairs $(v,M^v_{cap})$; (3) memory efficiency ratios for each vehicle $(M^v_{cap}/mem_v)$; (4) computation time $t^{cmp}_v$ and communication time $t^{com}_v$ for each vehicle under current model partition; and (5) the execution path $p$ derived from the cluster DAG $(V,E)$.
\par
The action space involves two coupled decisions: model partition assignment, selecting the target vehicle and partition size $(v, M^v_{cap})$, and vehicle scheduling, determining execution order within the DAG's constraints. 
Together, these decisions are combine  to form pipeline configurations that balance resource allocation and computational efficiency across the network.
The reward function balances performance with constraint satisfaction via weighted summation, and we have
\begin{align}
r \leftarrow & w_1(-(t^v_{cmp} + t^v_{com})) + w_2\mathbbm{1}(M^v_{cap} \leq mem_v) \nonumber \\
&+ w_3\mathbbm{1}(M^v_{cap} \cap M^u_{cap} = \emptyset) + w_4\mathbbm{1}(\text{DAG valid}).
\end{align}
For terminal states, we incorporate path execution time $r \leftarrow r - t_{path}(p_j,P)$. This guides the DQN toward efficient, feasible pipeline configurations satisfying both performance and operational constraints.
\subsection{\textbf{Quick Recovery for Fast Adaption}}
While SWIFT generates efficient pipelines, the dynamic nature of VFL presents significant challenges to FLAD's robustness and performance. 
The high mobility of vehicles introduces two critical vulnerabilities to the training process. 
First, when a vehicle departs or disconnects unexpectedly, the cluster may lose critically trained weights, potentially causing pipeline failure. Second, network anomalies can create pipeline blockages when vehicles disconnect temporarily, necessitating pipeline reconstruction. 
\par
Traditional approaches that rely on model aggregation at a central coordinator followed by complete replanning, relaunching and redistribution introduce substantial delays that impact training efficiency.
The proposed dynamic quick recovery mechanism addresses these challenges through two complementary modules that handle resource fluctuations. The first module employs preventive pipeline template-based fault tolerance, while the second leverages edge-aided backup and recovery strategies.
\par
{\textbf{Preventive Pipeline Template-based Fault Tolerance.}}
FLAD enhances training resilience through proactive pipeline generation, significantly reducing computation overhead during runtime. 
This approach pre-generates pipeline configurations for potential stage disconnections, ensuring rapid recovery when non-initial pipeline stages fail. 
Cluster maintains an essential pipeline set $\mathcal{P}^e$, assigning each vehicle an efficient pipeline profile $(p_v,\mathcal{P}_v)$.
For each vehicle $v \in V$, template generation follows a two-step process: (1) identification of viable cluster subsets $Clu$ containing vehicle $v$ with sufficient resources to handle model capacity $M_{cap}$, and (2) generation of refined pipeline templates using DQN-based pipeline generation method. 
This process runs concurrently with the essential pipeline set initialization and training, enabling immediate template deployment during failures without replanning and relaunch delays.
\par
{\textbf{Edge-aided Backup and Recovery.}}
The edge server (master node) maintains periodic model state backups every $e$ epochs under the active pipeline template $(p_v,\mathcal{P}_v)$, balancing backup frequency with network stability conditions.
When failures occur, the edge server executes a streamlined recovery process by deploying pre-generated templates $(p_v^{\prime},\mathcal{P}_v^{\prime})$ and comparing the difference between original and new configurations. 
This targeted approach distributes only modified model partitions to corresponding vehicles, minimizing communication overhead and accelerating recovery.
\section{Implementation Details}
We implement a complete FLAD prototype system in Python, leveraging PyTorch and PyTorch DDP~\cite{paszke2017automatic,li2020pytorch} as the foundation for our distributed training framework.

\subsection{Implementation of FHDP}
\begin{figure}[!t]
		\centering
		\subfloat[The AD vehicle with Jetson.]{
			\includegraphics[width=0.5\linewidth]{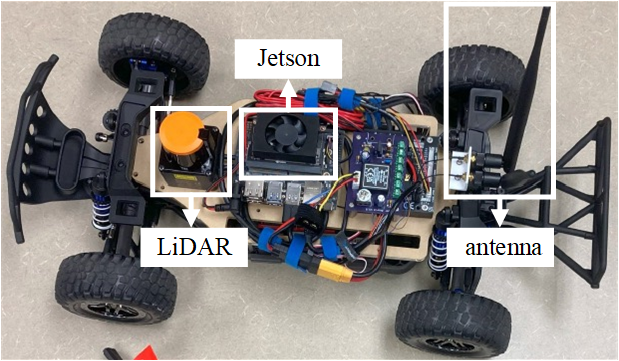}
			\label{fig_F1tenth_case1}}
        \hfil
		\subfloat[The cluster of FHDP.]{
			\includegraphics[width=0.375\linewidth]{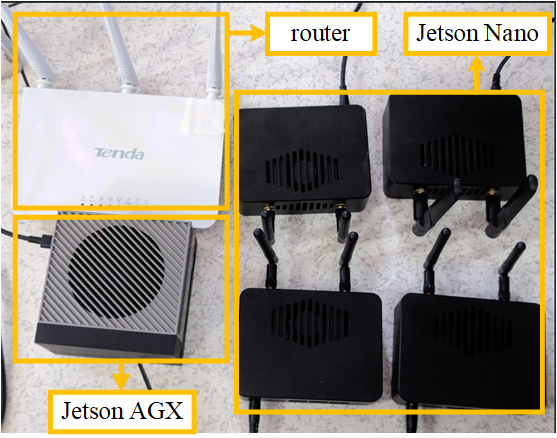}
			\label{fig_F1tenth_case4}}
		\caption{FHDP testbed.}
		\label{fig_protype}
	\end{figure}

As shown in Fig.\ref{fig_protype}, our FHDP prototype utilizes a heterogeneous testbed of Jetson devices (four Nano, two NX, one AGX), representing diverse vehicular computational capabilities~\cite{10749837}.
Table \ref{tab:experimental_devices} summarizes their hardware specifications, showing notable variations in memory and computation performance.
We implement intra-cluster pipeline parallelism using the Pippy library~\cite{pippy2022} and inter-cluster FL with Plato~\cite{tl-systemplato_2025}.
\par
The edge server hosts the master node, which orchestrates pipeline execution and monitors cluster health. Each participating device initializes a dedicated rank worker with an RPC backend for inter-device communication. The master node analyzes model computation graphs through PyTorch's execution tracing~\cite{paszke2017automatic} to generate efficient pipeline configurations.
\par
We address three significant implementation challenges:
\paragraph{Unified CPU/GPU Memory Management.} Jetson's architecture unifies CPU and GPU memory, creating resource contention between system operations and model training. 
Unlike devices with a separate GPU, memory overcommitment not only affects training stability, but also critical system functions.
We developed an active memory management mechanism that periodically reclaims training resources, significantly improving system robustness without compromising training efficiency.
\paragraph{Dynamic Model Partitioning.} PyTorch's \textit{torch.fx} provides intermediate representation (IR) capabilities for computation graph analysis, operating between operator layers and forward functions. While this framework allows flexible graph modification through its doubly linked list structure—supporting submodules, external functions, and instance methods—it fundamentally struggles with elements critical to AD models: dynamic control flows, conditional branches, and multimodal inputs. When encountering such complexity, symbolic tracing fails immediately without graceful fallbacks. We address this limitation through a decorator-based approach that strategically relocates dynamic control instructions from GPU to CPU, enabling flexible model partitioning while maintaining execution semantics across distributed vehicular resources.
\paragraph{Fault-Tolerant Training.} Conventional failure recovery mechanisms simply restart training, disregarding partial progress and healthy components. Through fine-grained analysis of FHDP's pipeline structure, we developed a dynamic recovery system that preserves operational stages on healthy devices while reallocating failed components. This approach minimizes recovery overhead and maintains training continuity during device failures.
\begin{table}
   \centering
   \caption{Hardware Specifications of Prototype Devices}
   \begin{tabular}{lcc}
   \toprule
   \textbf{Device} & \textbf{Memory} & \textbf{Compute Performance} \\
   \midrule
   Jetson NX &  8GB & 0.404 TFLOPS \\
   Jetson Nano &  8GB & 0.472 TFLOPS \\
   Jetson AGX &  32GB & 3.85 TFLOPS \\
   \bottomrule
   \end{tabular}
   \label{tab:experimental_devices}
\end{table}
\subsection{Implementation of LLM with CELLAdapt}
In the CELLAdapt framework, we designed an LLM-based pipeline that adapts cloud-based LLMs to edge environments while preserving AD-specific knowledge. Our implementation includes a knowledge distillation part and a fine tuning part, which employs the HuggingFace Transformers library to deploy a LLaMA-7B and LLaMA-3B foundation model, enhanced with LoRA to minimize memory requirements during adaptation. 
\par
The knowledge distillation process abstracts the AD knowledge from cloud LLMs to AD-LLM in the cloud. 
The edge server also runs the distillation to compress AD-LLM to a smaller ADM, where LLaMA-7B is used for AD-LLM and LLaMA-3B for ADM. The $L_1-norm$ loss is adopted to align the outputs (i.e., waypoints) of the teacher and student models. On the other hand, the AD-LLM is fine tuned by the client features to improve the adaptation for data in the specific area. The entire LLM pipeline operates asynchronously with the FHDP framework, allowing continuous model refinement without disrupting vehicular operations. This architecture enables efficient and accurate decision-making for autonomous vehicles. 

\section{Evaluation}
\subsection{Experiment Settings}
We evaluated FLAD using a heterogeneous computing setup with two high-performance servers: one with three NVIDIA V100-32GB GPUs and another with a single NVIDIA A100-80GB GPU. This configuration simulated both cloud-tier and edge computing environments.
\par
We evaluated FLAD using synthetic driving data generated with CARLA simulator following \cite{shao2024lmdrive}. The dataset contains 5-15 second clips across diverse urban scenarios, including four-camera RGB images, LiDAR point clouds, and sensor telemetry (GPS, weather, vehicle data).
To simulate realistic federated learning conditions, we distributed data across 50 virtual vehicles with controlled non-IID characteristics based on CARLA town environments. Multimodal sensory data supported vision encoder training in FHDP, while derived textual content enabled AD-LLM adaptation in CELLAdapt.
Our evaluation framework assessed two dimensions: system efficiency (memory usage, throughput) and driving performance (Route Completion, Infraction Score, and their product as Driving Score). We conducted comprehensive closed-loop testing across various CARLA environments, with federated learning experiments using 5 active clients per round, non-IID level 2, 5 local epochs, batch size 4, and dual learning rates (0.00075 general, 0.0003 backbone) with Adam optimization.
\subsection{Performance of SWIFT}

\begin{figure*}[!t]
  \centering
  \begin{minipage}{0.3\linewidth}
    \centering
    \setlength{\abovecaptionskip}{0.cm}
    \includegraphics[width=0.9\linewidth]{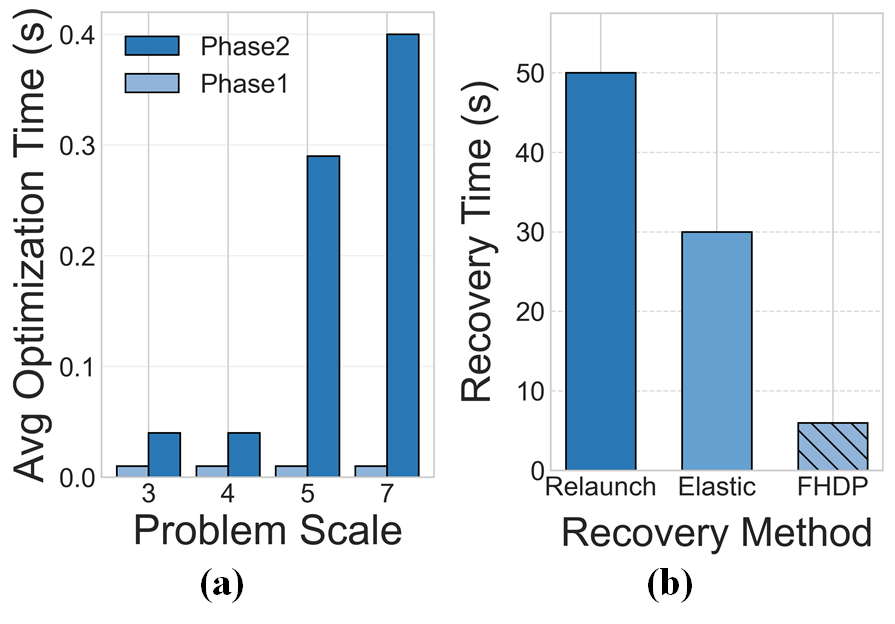}
    \caption{
    Averaging (a) optimization time and (b) recovery time of SWIFT.}
    \label{fig_swift_time}
  \end{minipage}%
  \hfil
  \begin{minipage}{0.3\linewidth}
    \centering
    \setlength{\abovecaptionskip}{0.cm}
    \includegraphics[width=0.9\linewidth]{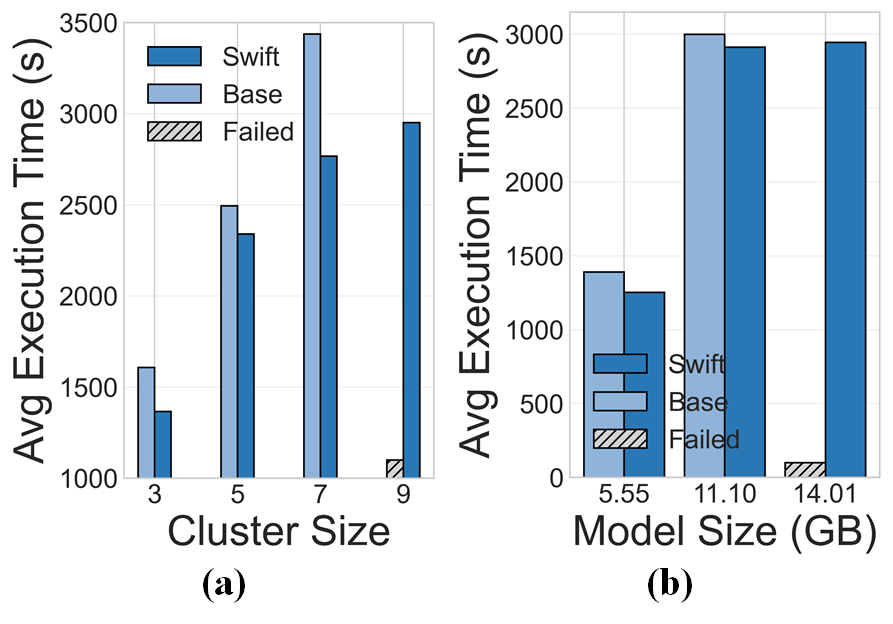}
    \label{fig_swift_case3}
    \caption{SWIFT pipeline execution time with (a) different cluster size and (b) different model size.}
    \label{fig_swift_performance}
  \end{minipage}
    \hfil
  \begin{minipage}{0.3\linewidth}
    \centering
    \setlength{\abovecaptionskip}{0.cm}
    \includegraphics[width=0.9\linewidth]{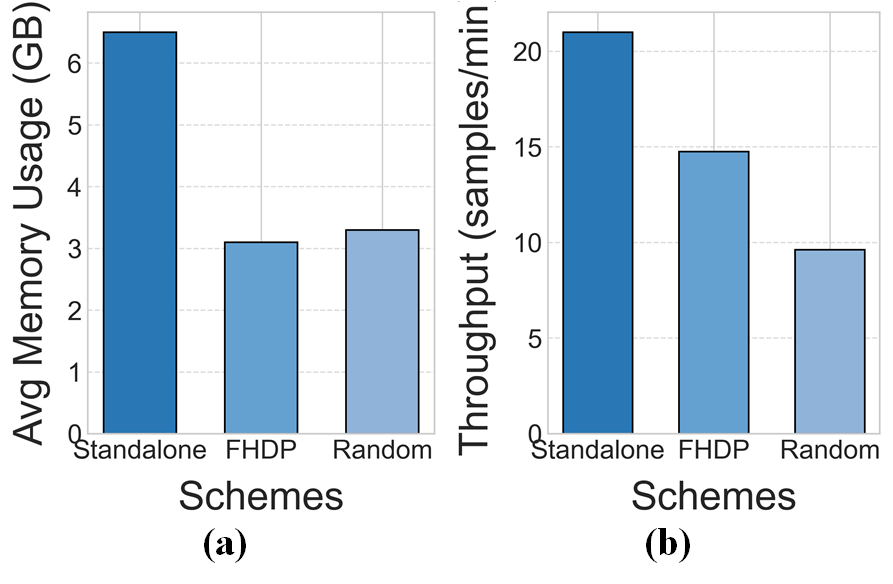}
    \caption{The (a) memory footprints and (b) throughout in training vision encoder.}
    \label{fig_FHDP_performance}
  \end{minipage}
\end{figure*}

As shown in Fig.~\ref{fig_swift_time}(a), Phase 1 of SWIFT runs significantly faster than the DQN-based Phase 2, maintaining stable optimization time and enabling immediate pipeline execution. 
This observation demonstrates that SWIFT's greedy-based initial phase enables fast start and flexible scheduling across heterogeneous vehicular resources.

In Fig.~\ref{fig_swift_performance}(a), SWIFT demonstrates consistent performance advantages over greedy matching, reducing execution time by 15\% at cluster size 5 and successfully handling cluster size 9 where the baseline fails completely. This improvement is due to SWIFT's joint optimization of coupled resources, as formulated in Eq.\eqref{eq:pipeline generation}, whereas the baseline's single-resource optimization approach produces suboptimal or infeasible solutions as complexity increases.

In Fig.~\ref{fig_swift_performance}(b), we fix the cluster size to 5 vehicles and evaluate SWIFT’s performance with increasing model sizes. It achieves a 9.9\% reduction at 5.55 GB (1252.81s vs. 1390.74s), a 2.9\% reduction at 11.10 GB (2911s vs. 2998.28s), and successfully executes at 14.01 GB, while the baseline fails to handle excessive resource demands.

These results demonstrate SWIFT’s superior efficiency, scalability, and robustness in handling large-scale pipeline training across varying cluster sizes and model complexities, making it well-suited for real-world deployment.
\subsection{Performance comparison of FHDP}
\begin{table*}[t]
    \caption{Network Communication Characteristics with Different Pipelines.}
    \centering
    \begin{tabular}{lrr|lrr}
    \toprule
    \multicolumn{3}{c}{\textbf{Random}} &\multicolumn{3}{c}{\textbf{FHDP}}\\
    \midrule
    \textbf{Metric} & \textbf{Stage 1} & \textbf{Stage 2}& \textbf{Metric} &\textbf{Stage 1} & \textbf{Stage 2} \\
    \midrule
    Training duration (s) &1301 &1296 &Training duration (s)&871 &875 \\
    Data size (MB) & 2440 & 2428 & Data size (MB)& 2455 & 2427\\
    Network throughput (Mbps) & 15 & 14 &Network throughput (Mbps) & 22 & 22 \\
    Average packet size (bytes) & 1121.23 & 1800 &Average packet size (bytes) & 1130 & 1939.7 \\
    Average packet rate (packets/s) &1672&1039&Average packet rate (packets/sec)&2492&1429\\
    Number of packets & 2176k & 1348k &Number of packets & 2171k & 1251k \\
    \bottomrule
    \end{tabular}
    \label{tab:network_characteristics}
\end{table*}
\begin{figure*}[ht]
    \centering
    \setlength{\abovecaptionskip}{-0.5cm}
    \includegraphics[width=\linewidth]{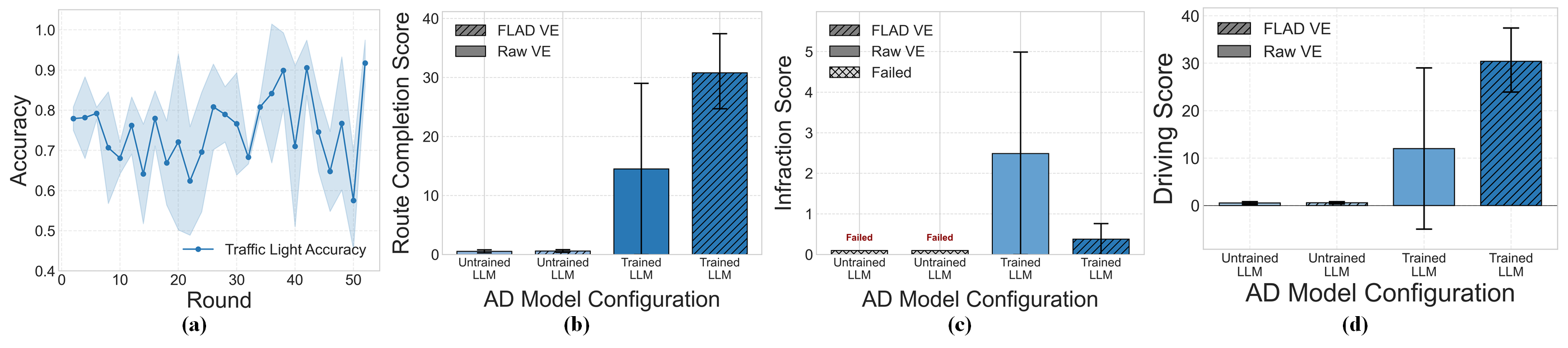}
    \label{fig_FLAD_case4}
    \caption{The performance of FHDP including (a) performance improvement of vision encoder by FLAD training, (b) route completion scores under different LLM-vision encoder configurations, (c) infraction scores under different LLM-vision encoder configurations, and (d) driving scores under different LLM-vision encoder configurations.}
    \label{fig_FLAD_performance}
\end{figure*}
As shown in Fig.~\ref{fig_swift_time}(b), the proposed fault-tolerance mechanism achieves the recovery time of 5 seconds, significantly outperforming relaunch and Elastic TorchRun, which require 50 and 30 seconds, respectively. This improvement is thanks to our mechanism preserving the original low-level communication stack, reassigning only stage IDs and corresponding submodules to vehicles, thus minimizing recovery overhead.

In Figs.~\ref{fig_FHDP_performance}(a) and (b), we validate FHDP’s performance on our testbed, which consists of two Jetson Nano devices and one Jetson AGX, simulating a heterogeneous cluster. We partition the model using SWIFT and compare FHDP against a random solution and a standalone solution as baselines, i.e. the model is trained in a single node with sufficient compute and memory resources, completely eliminating communication overhead.
\par
As shown in Fig.~\ref{fig_FHDP_performance}(a), FHDP achieves an average memory footprint of 3.1GB, lower than the random solution’s 3.3GB. This reduction results from FHDP selecting a more efficient partitioning strategy that minimizes both model size and the amount of activation data transmitted.
In Fig.~\ref{fig_FHDP_performance}(b), FHDP achieves a 40\% higher throughput than the random solution and reaches 75\% of the standalone solution’s throughput. 
\par
To further evaluate FHDP's effectiveness, we implemented a testbed with a Jetson Nano (stage 1) and Jetson AGX (stage 2), comparing random splitting versus our SWIFT-generated configuration. Table~\ref{tab:network_characteristics} shows that despite similar data volumes, FHDP achieves 1.47$\times$ higher network throughput and 33\% shorter training duration. This improvement stems from SWIFT's intelligent model partitioning, which optimizes heterogeneous resource utilization and maintains higher packet rates (2492 vs. 1672 packets/s) under identical hardware constraints.

\subsection{Performance of FLAD with CELLAdapt}

\begin{figure}[ht]
\centering
\subfloat[Pedestrian avoidance scenario]{
    \includegraphics[width=0.44\linewidth]{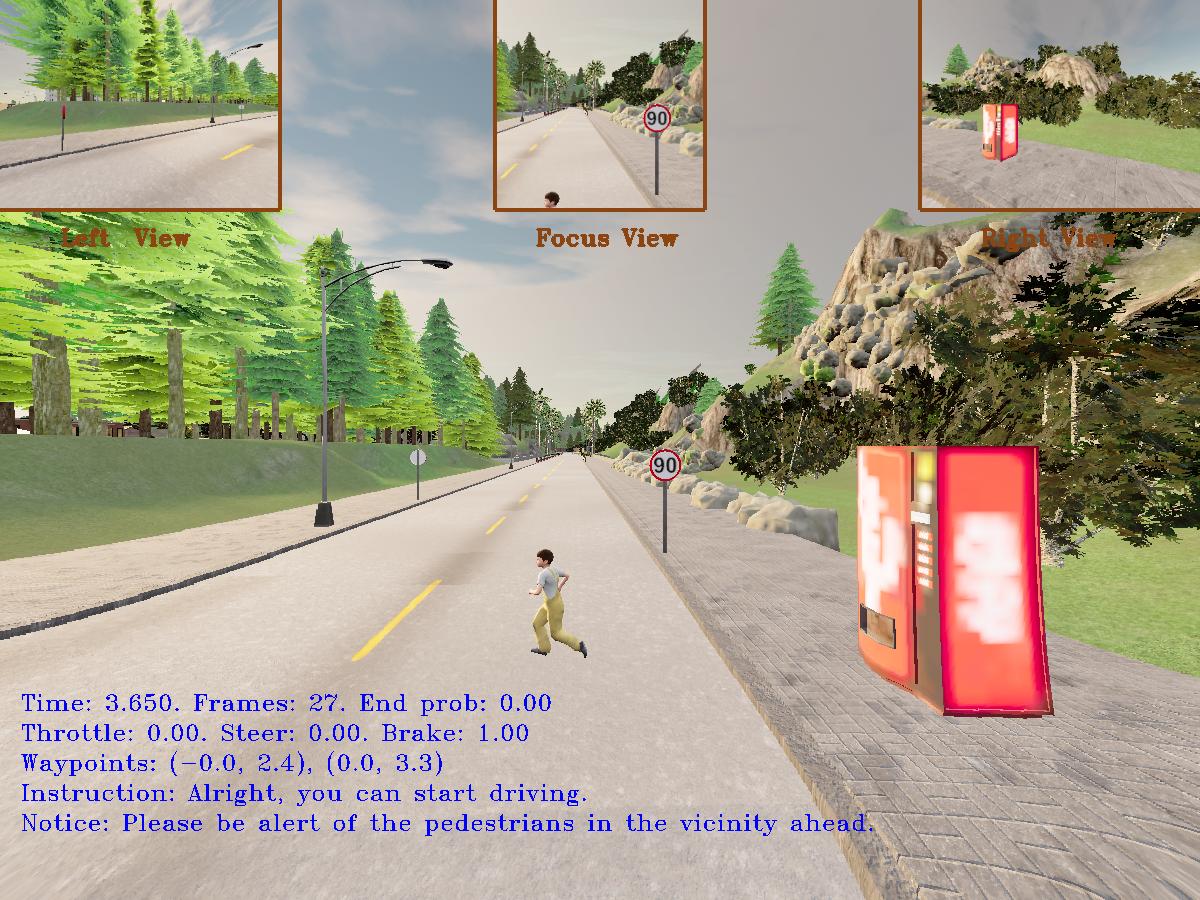}
    \label{fig_Carla_case1}}
\hfil
\subfloat[Oncoming traffic navigation]{
    \includegraphics[width=0.44\linewidth]{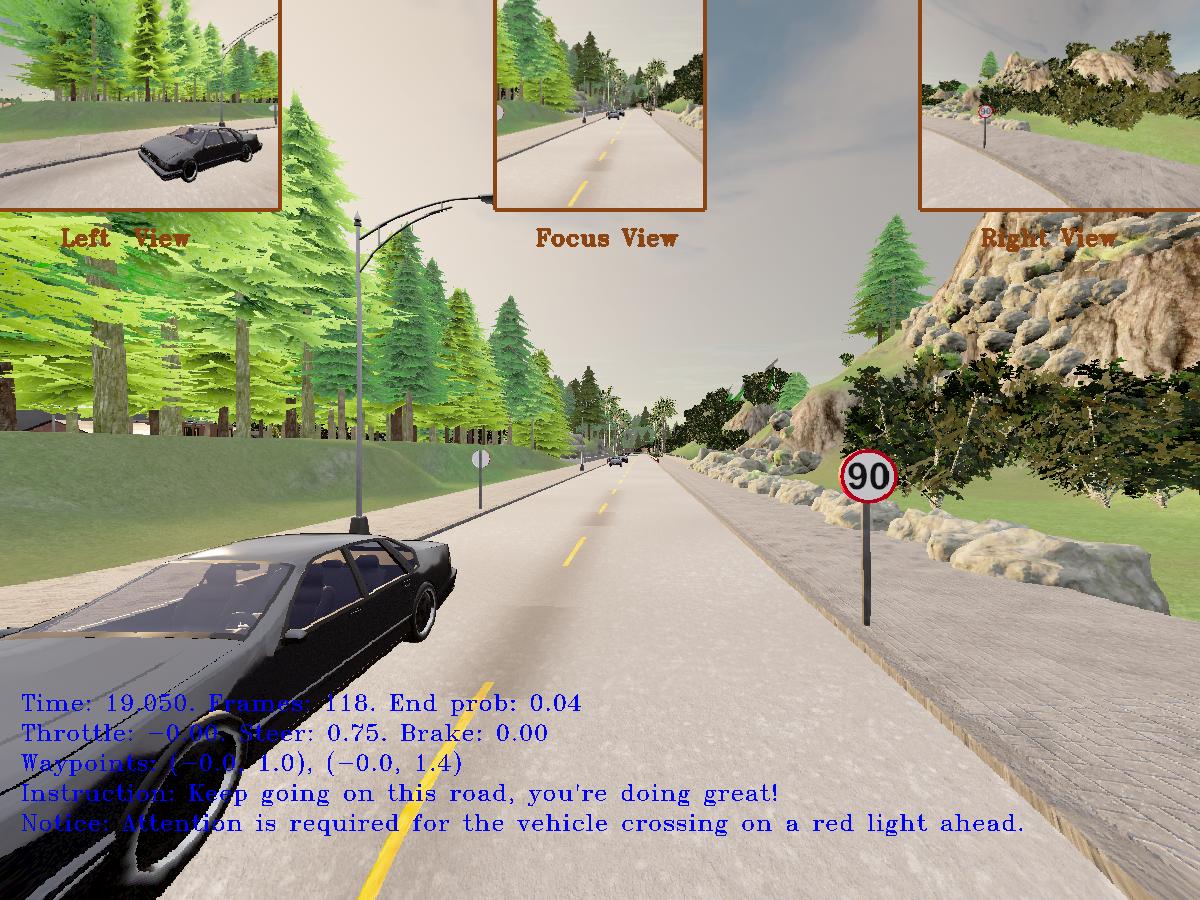}
    \label{fig_Carla_case2}}
   \hfil
\subfloat[Pre-lane change decision]{
    \includegraphics[width=0.44\linewidth]{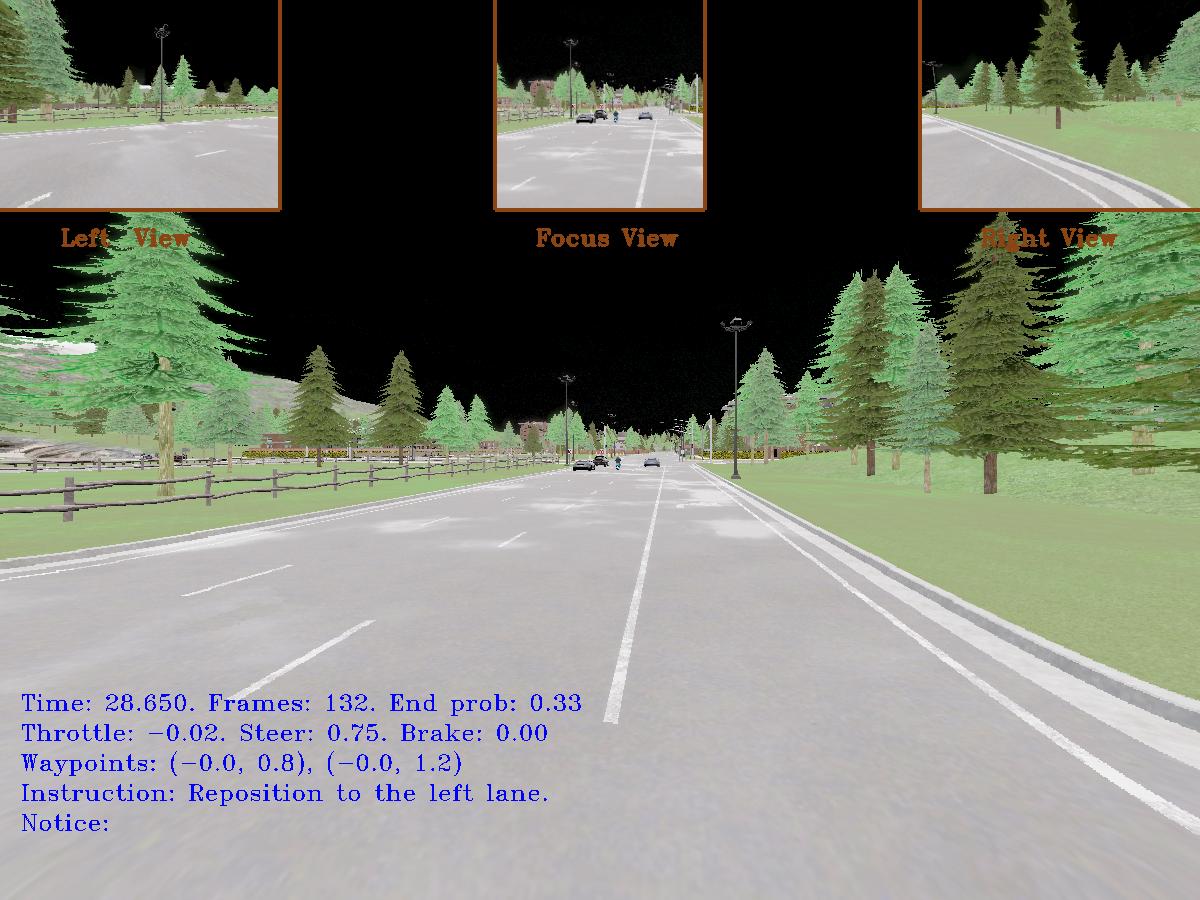}
    \label{fig_Carla_case3}}
   \hfil
\subfloat[Post-lane change execution]{
    \includegraphics[width=0.44\linewidth]{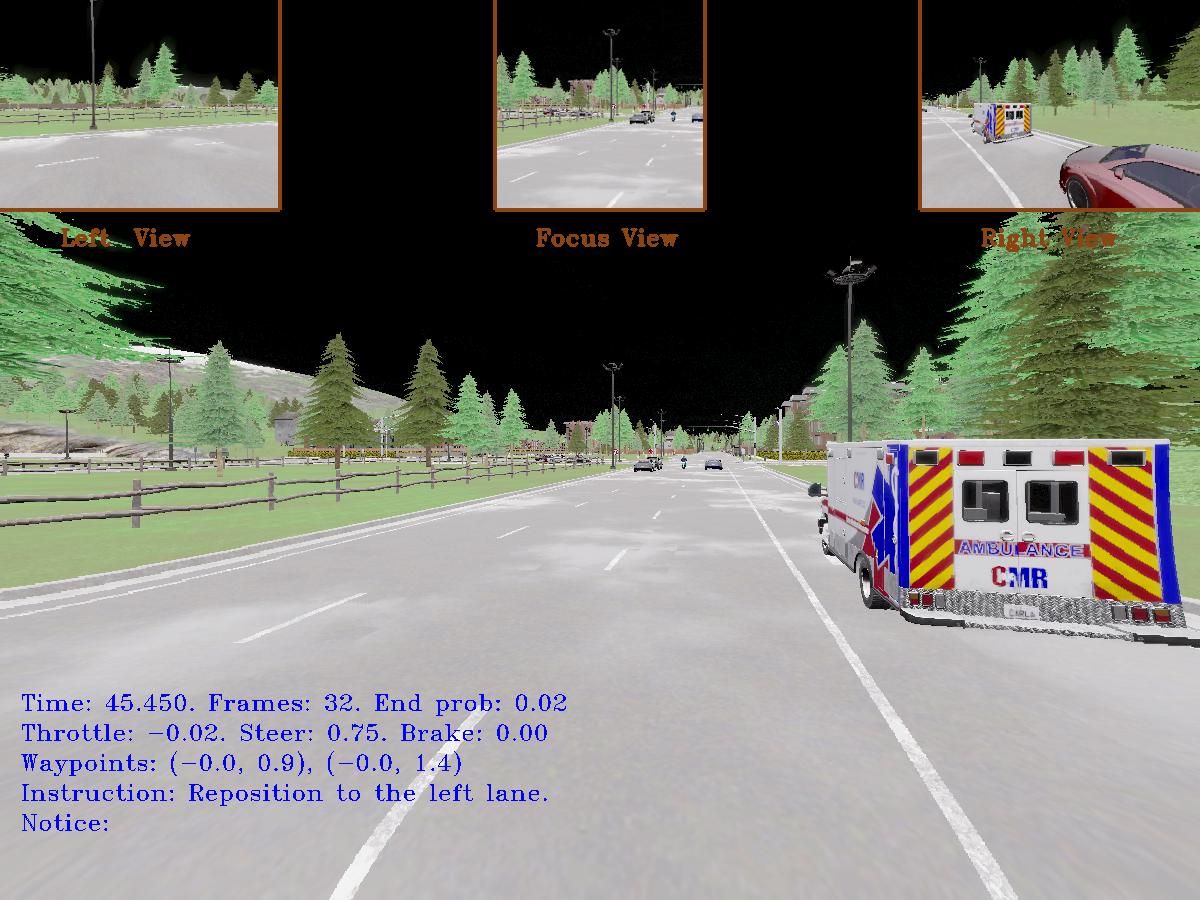}
    \label{fig_Carla_case4}}
\caption{FLAD autonomous driving decision examples across complex traffic scenarios.}
\label{fig_Carla_example}
\end{figure}
Fig.~\ref{fig_FLAD_performance}(a) demonstrates the effectiveness of FLAD in improving traffic light prediction accuracy from 79.9\% to 92.66\% despite non-i.i.d. data distribution. 
Specifically, these improvements are achieved by training  previously centrally trained encoders in FLAD.
The improved performance is attributed to the personalization of global model trained with the data from specific towns in our settings.
As shown in Figs.~\ref{fig_FLAD_performance}(b)-(d), CARLA evaluations show that FLAD-trained encoders paired with domain-specific LLMs achieve 15\% higher Route Completion and 16\% higher Driving Scores with fewer infractions, confirming FLAD's effectiveness for region-specific AD despite LLM quality remaining the primary performance determinant.
\par
Fig.\ref{fig_Carla_example} demonstrates FLAD's performance across challenging scenarios: pedestrian avoidance with appropriate deceleration as shown in Fig.\ref{fig_Carla_example}\subref{fig_Carla_case1}, opposite-direction traffic navigation with robust lane positioning as shown in Fig.\ref{fig_Carla_example}\subref{fig_Carla_case2}, and complete lane-change execution from intention to completion as shown in Figs.\ref{fig_Carla_example}\subref{fig_Carla_case3}-\subref{fig_Carla_case4}. These results confirm FLAD's capability to handle complex driving situations requiring integrated perception, decision-making, and control.
\begin{figure}[ht]
    \centering
    \setlength{\abovecaptionskip}{0.cm}
    \includegraphics[width=\linewidth]{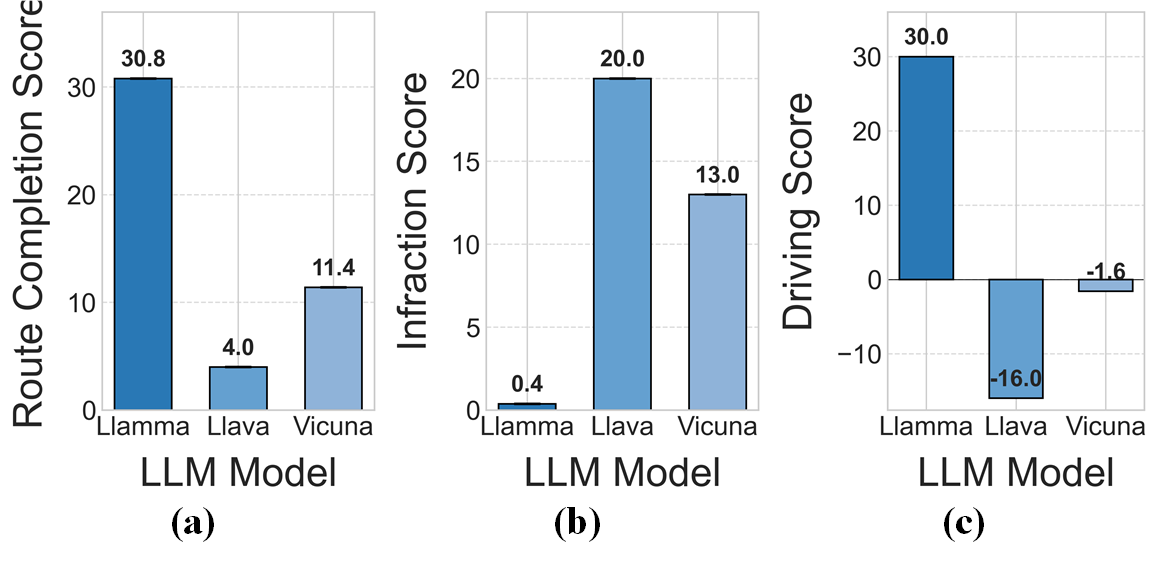}
    \caption{The FLAD performance with different LLMs including (a) route completion scores, (b) infraction scores, and (c) driving scores.}
    \label{fig_LLM_performance}
\end{figure}
\par
Fig.~\ref{fig_LLM_performance} demonstrates that FLAD with LLaMa-7B significantly outperforms other LLM configurations (LLaVA-7B and Vicuna), achieving superior Route Completion (30.8), Infraction Score (0.38), and Driving Score (30). These results confirm that FLAD produces more accurate, personalized autonomous driving models capable of handling complex driving tasks effectively.
\section{Related Work}
\textbf{End-to-End Autonomous Driving.}~Recent advancements in end-to-end AD leverage vision-language models, multimodal fusion, and unified architectures. DriveAdapter \cite{jia2023driveadapter} aligns perception and planning features, while DriveLM Agent \cite{sima2023drivelm} incorporates vision-language reasoning for making decision. Talk2BEV~\cite{choudhary2023talk2bev} enhances spatial understanding via BEV maps. Multimodal fusion approaches, such as LMDrive~\cite{shao2024lmdrive}, integrate sensor data with natural language navigation. DriveDreamer~\cite{wang2024drivedreamer} introduces a world model trained on real-world scenarios, and UniAD~\cite{hu2023planning} unifies feature abstractions for holistic interaction modeling. While these approaches enhance generalization, they rely on centralized data centers, raising privacy concerns and limiting adaptability. Our work shifts toward FL, enabling privacy-preserving, region-specific adaptation while improving resource efficiency.

\textbf{Parallelism in Datacenter.}~Scalable model training relies on various parallelism strategies in datacenters. ZeRO~\cite{rajbhandari2020zero} reduces memory redundancy for trillion-parameter models, while Megatron~\cite{shoeybi2019megatron} improves intra-layer parallelism. Pipeline parallelism frameworks, including TensorPipe~\cite{huang2019gpipe} and DAPPLE~\cite{fan2021dapple}, optimize layer-based training, while hybrid approaches like nnScaler~\cite{lin2024nnscaler} balance resource utilization. Oobleck~\cite{jang2023oobleck} enhances fault tolerance in distributed training. However, these techniques assume stable, high-speed network connectivity, making them unsuitable for vehicular edge environments characterized by bandwidth variability and dynamic topologies. FLAD adapts parallelism techniques for decentralized, resource-constrained environments while preserving data privacy vian FL.

\textbf{Machine Learning Scheduling.}~Efficient ML scheduling improves resource utilization in constrained environments. Asteroid~\cite{ye2024asteroid} accelerates fault-tolerant pipeline training, while EdgePipe~\cite{yoon2021edgepipe} optimizes execution under fluctuating network conditions. StellaTrain~\cite{lim2024accelerating} and FlexNN~\cite{li2024flexnn} enhance network-aware scheduling, and AutoFed~\cite{zheng2023autofed} addresses multimodal sensor heterogeneity in FL. Alpa~\cite{zheng2022alpa} unifies parallelism strategies across heterogeneous devices. However, existing solutions do not fully account for extreme mobility, network instability, and LLM-specific computational demands in vehicular environments. FLAD bridges this gap by integrating FL, pipeline parallelism, and cloud-edge-vehicle collaboration to enable efficient, privacy-preserving AD model training.

\section{Conclusion}
In this paper, we proposed FLAD, a heterogeneity-aware FL framework for LLM-based autonomous driving. FLAD optimizes model training, communication efficiency, and resource utilization through (1) a cloud-edge-vehicle architecture, (2) intelligent clustering, pipelining, and communication scheduling, and (3) knowledge distillation for personalization. Experimental results demonstrate that FLAD enhances driving performance while efficiently leveraging distributed vehicular resources, making it a promising solution for scalable and privacy-preserving AD deployment.
\par
Despite these advancements, we identified a bottleneck in FLAD due to the large data size and slow data loading process. Additionally, frequent communication between pipeline stages results in significant wireless network overhead.
Future work will focus on designing more efficient models to reduce training costs, optimizing the data loading process, and compressing communication overhead to further enhance training efficiency.
\par
In summary, FLAD presents new opportunities to leverage the limited computing and storage resources of hardware-constrained end devices for collaborative training of effective models to locally process vast volumes of multi-modal sensing information. With the assistance of edge and cloud computing, these devices can utilize more complex large models to handle autonomous tasks that would otherwise be impossible. The applications of FLAD can be extended to assist autonomous tasks for drones, UAVs, robots, and more, making it a promising approach for the future of embodied AI.

\bibliographystyle{ACM-Reference-Format}
\bibliography{ref}

\appendix

\end{document}